\documentclass{article}
\usepackage{spconf,amsmath,graphicx,hyperref}
\usepackage{amssymb}
\usepackage{booktabs}
\usepackage{array}
\usepackage{multirow}
\usepackage{svg}
\usepackage{tabularx}
\usepackage{xcolor}

\title{ERPBench: A State-Grounded Evaluation Paradigm for Computer-Use Agents in Enterprise Software \thanks{\copyright{} 2026 A\lowercase{ccenture}. A\lowercase{ll rights reserved}.}}
\name{\begin{tabular}[t]{@{}c@{}}
  Kratika Bhagtani$^{*}$, Kusha Sridhar$^{*}$, Maziyar Baran Pouyan, \\
  Yuying Zhao, Eugene Siow
\end{tabular}\thanks{$^{*}$These authors contributed equally.}}
\address{Agentic AI Center of Excellence, Accenture \\
  {\small\texttt{\{kratika.bhagtani, k.sridhara.murthy, maziyar.baran.pouyan,}} \\
  {\small\texttt{yuying.d.zhao, eugene.siow\}@accenture.com}}}
\newcolumntype{C}{>{\centering\arraybackslash}X}
\begin{document}
%
\maketitle
\begin{abstract}
Computer-use agents that operate through screenshots and simulated actions are advancing rapidly, yet their evaluation remains anchored to general desktop and web tasks. 
Enterprise Resource Planning systems run the finance, procurement, inventory, and customer operations of organizations worldwide, and pose distinct challenges for computer-use agents: dense interfaces, coordinated multi-step interactions, and errors that alter persistent business records rather than surfacing on screen.
Existing enterprise computer-use benchmarks rely on proprietary platforms or on simulated approximations of such software.
We introduce ERPBench, a benchmark that evaluates screenshot-only agents on a live and reproducible system and scores each task against ground-truth values in its database. 
Beyond the benchmark, we present a production-grade harness that gates agent actions behind human approval for safe deployment.
Evaluating six closed and open-source agents, we demonstrate that strong general performance does not transfer to enterprise reliability.
Even when an agent reaches the right form and saves it, the stored record is often wrong: some agents save in up to 85\% of runs but write the correct value in as few as 3\%.
We further characterize failure modes specific to enterprise workflows.
\end{abstract}
\begin{keywords}
Computer-use agents, agent safety, enterprise software, human-in-the-loop, ERP
\end{keywords}

\section{Introduction}
\label{sec:intro}
\vspace{-5pt}
Advances in multimodal large language models (LLMs) have enabled \emph{computer-use agents} (CUAs) that operate graphical user interfaces (GUIs) directly, through screenshots and simulated mouse and keyboard actions~\cite{uitars2025, anthropic2024computeruse, openai2025cua, opencua2025}.
On general desktop and web benchmarks such as OSWorld~\cite{osworld2024}, Windows Agent Arena (WAA)~\cite{waa2024}, and WebArena~\cite{zhou2024webarena}, 
CUAs have improved rapidly (an example of a typical task is editing a spreadsheet in LibreOffice), suggesting that general GUI interaction is becoming tractable.
This progress does not necessarily translate to enterprise software.
Business systems such as Enterprise Resource Planning (ERP) and Customer Relationship Management (CRM), spanning finance, procurement, and inventory, differ fundamentally from consumer applications in both workflow structure and failure cost.
Here a visually plausible interaction is not enough: an agent may navigate to the correct screen, click the expected controls, and observe a success confirmation, yet leave the business record unchanged or wrong. The confirmation reflects that the interface accepted an action, not that the intended value reached the database: the value entered on screen may never have been bound to the field it appears to fill, so what persists is stale, empty, or wrong.
Such failures can be invisible from screenshots alone, yet they propagate into downstream operational and financial processes.
Existing GUI benchmarks primarily target general desktop and web environments~\cite{osworld2024, waa2024, zhou2024webarena}, while recent enterprise benchmarks use proprietary platforms or simulated enterprise environments~\cite{scuba2025, crmarena2025, cristescu2025ui}.
Neither decomposes a run into stages, so a failure to reach the form is scored like one where the agent saved and still stored the wrong value.
These limitations motivate an evaluation paradigm that measures business state and localizes failures within a workflow.

We call this paradigm \emph{state-grounded evaluation}, and instantiate it in \textbf{ERPBench}.  
ERPBench is built on a live, self-hosted instance of ERPNext~\cite{erpnext}, an open-source ERP spanning finance, procurement, inventory, and customer management. 
Agents operate from screenshots alone, and each task is graded against ground-truth database fields, extending execution-based state verification~\cite{osworld2024, zhou2024webarena} to business records. 
We impose the screenshot-only constraint because enterprise software is routinely reached through remote-desktop and Virtual Desktop Infrastructure (VDI) sessions that expose no Document Object Model (DOM), accessibility, or Application Programming Interface (API) hooks: pixel input is the only channel that generalizes across such deployments.
Field-level grading exposes enterprise-specific failure modes that metrics scored at the task level can conflate, including agents trapped in unrecoverable interaction loops.

\begin{table*}
\centering
\footnotesize
\setlength{\tabcolsep}{3pt}
\caption{Comparison with existing CUA benchmarks.
ITSM~=~IT Service Management;
CRM~=~Customer Relationship Management; ERP~=~Enterprise Resource Planning;
A11y~=~Accessibility tree; DOM~=~Document Object Model;
API~=~Application Programming Interface.
$\sim$~=~partial (SCUBA: difficulty labels, milestone rewards;
not structural tiers or fixed-stage grading).
\emph{Open / reproducible} = the environment is self-hostable without a
proprietary account or license.
A \checkmark\ under \emph{Needs structured access} is a limitation, not a
feature: such inputs ease grounding but assume A11y/DOM/API hooks that
legacy, pixel-only systems do not expose.
No prior benchmark satisfies every row; ERPBench is the only one to combine
all of them.}
\label{tab:comparison}
\begin{tabularx}{\textwidth}{@{} l *{7}{C} @{}}
\toprule
\textbf{Property}
  & \textbf{OSWorld} & \textbf{WAA} & \textbf{WorkArena}
  & \textbf{SCUBA} & \textbf{CRMArena-Pro} & \textbf{UI-CUBE} & \textbf{ERPBench (ours)} \\
\midrule
Domain
  & General & General & Enterprise
  & Enterprise & Enterprise & Enterprise & Enterprise \\
Subdomain
  & OS & OS & ITSM
  & CRM & CRM & Mixed & ERP \\
Platform
  & Ubuntu & Windows & ServiceNow
  & Salesforce & Salesforce & Web mocks & ERPNext \\
Distinct tasks
  & 369 & 154 & 33 & 60 & 19 & 226 & 30 (150 runs) \\
\midrule
Pixel-only input
  & $\times$ & $\times$ & $\times$
  & \checkmark & $\times$ & $\times$ & \checkmark \\
Needs structured access
  & \checkmark (A11y) & \checkmark (A11y) & \checkmark (A11y{+}DOM)
  & $\times$ & \checkmark (Text/API) & \checkmark (DOM) & $\times$ \\
Live, non-simulated app
  & \checkmark & \checkmark & \checkmark
  & \checkmark & \checkmark & $\times$ & \checkmark \\
Open / reproducible
  & \checkmark & $\times$ & $\times$
  & $\times$ & $\times$ & \checkmark & \checkmark \\
Database-verified ground truth
  & $\times$ & $\times$ & \checkmark
  & \checkmark & \checkmark & $\times$ & \checkmark \\
Human baseline
  & \checkmark & \checkmark & $\times$
  & $\times$ & $\times$ & \checkmark & \checkmark \\

\midrule
Task complexity tiers
  & $\times$ & $\times$ & $\times$
  & $\sim$ & $\times$ & \checkmark & \checkmark \\
Stage-wise grading
  & $\times$ & $\times$ & $\times$
  & $\sim$ & $\times$ & $\times$ & \checkmark \\
Formal failure taxonomy
  & $\times$ & $\times$ & $\times$
  & $\times$ & $\times$ & $\times$ & \checkmark \\
\bottomrule
\end{tabularx}

\end{table*}

ERPBench runs on a production-grade harness that we present. 
In deployment, it gates irreversible agent actions behind human approval for safe enterprise use~\cite{anthropic2024computeruse}.
For benchmarking, the same harness runs autonomously, with an auto-approver in place of the human reviewer.
We evaluate six closed and open-source CUAs~\cite{anthropic2024computeruse, uitars2025, opencua2025, qwen3vl2025, hai2025holo3modelfamily} on tasks spanning single-field form edits, multi-field business record creation, and multi-screen chained workflows.
We demonstrate that high apparent GUI competence does not imply enterprise reliability: agents that navigate successfully often fail at the interaction or commit stage, and even the strongest proprietary model exhibits enterprise-specific failures that existing methodologies do not capture.
Our contributions are threefold: 
(1) \textbf{ERPBench}, the first benchmark to operationalize pixel-only, state-grounded evaluation for enterprise CUAs on a live, openly reproducible enterprise system, grading screenshot-only agents at the database-field level on an open-source ERP~\cite{erpnext}.
(2) \textbf{A production-grade harness} with human-in-the-loop safety gating, run autonomously to benchmark the same system deployed in production.
(3) \textbf{The first systematic analysis of enterprise-specific CUA failures}, showing that interaction-level evaluation overstates enterprise reliability, and localizing the stage at which agents break down.

\section{Related Work}
\label{sec:related}
\vspace{-5pt}


Computer-use agent evaluation has developed along four threads: general GUI benchmarks, enterprise computer-use evaluation, agent safety, and computer-use agent systems.

\noindent\textbf{General GUI benchmarks:}
Recent benchmarks have established standardized environments for evaluating CUAs across desktop and web applications. OSWorld~\cite{osworld2024} evaluates multimodal agents on open-ended tasks in real desktop environments, while Windows Agent Arena~\cite{waa2024} extends evaluation to Windows. WebArena~\cite{zhou2024webarena} provides realistic web environments and evaluates task completion against the underlying environment state rather than the rendered page. Collectively, these benchmarks establish interaction and execution success as the dominant evaluation paradigm for general-purpose GUI agents, but do not model the domain-specific business state characteristic of enterprise applications. ERPBench builds on the execution-based verification principle of WebArena~\cite{zhou2024webarena} but extends it to relational business records verified at the database-field level.

\noindent\textbf{Enterprise Computer-Use Evaluation:}
Several recent benchmarks extend computer-use evaluation to enterprise workflows. WorkArena~\cite{workarena2024} evaluates web agents on knowledge work tasks in ServiceNow, while Salesforce Computer Use Benchmark (SCUBA)~\cite{scuba2025} evaluates computer-use agents on Salesforce CRM workflows using application APIs for binary and milestone-based grading. CRMArena-Pro~\cite{crmarena2025} evaluates LLM agents on diverse CRM workflows, but uses text-based interaction rather than screenshot-driven GUI control. 
EntWorld~\cite{mo2026entworld} evaluates enterprise GUI agents across six dockerized open-source business applications with SQL-based deterministic verification and schema-driven task generation. However, EntWorld exposes an accessibility tree alongside screenshots and does not decompose runs into fixed stages or define a named failure taxonomy. 
UI-CUBE~\cite{cristescu2025ui} is particularly close to our work, arguing, as we do, that task-level accuracy alone is insufficient to assess enterprise readiness and introducing operational-reliability evaluation across enterprise workflows. However, UI-CUBE evaluates agents on simulated web applications, exposes DOM information alongside screenshots, and verifies outcomes against application-state snapshots. ERPBench instead evaluates screenshot-only agents on a live, self-hosted ERP deployment and grounds correctness in database fields. Moreover, ERPBench decomposes execution into fixed stages and introduces a named failure taxonomy to localize where enterprise agents fail.



\noindent\textbf{Agent Safety and Human Oversight:}
Deploying CUAs in enterprise systems also raises safety concerns because GUI actions may modify or irreversibly commit business records. Prior work has studied \emph{human-in-the-loop} (HITL) approval, automated intent verification, and agent safety awareness as mechanisms for controlling risky actions~\cite{turan2026oversightcapacitycalibratingagent,wang2026reframingllmagentsecurity,chen2026lpsbenchbenchmarkingsafetyawareness,zhang2026invisibleinkthreatsadversarial}. These works primarily evaluate safety mechanisms or agent-side risk awareness. ERPBench instead incorporates a risk-tiered approval gate into a production-oriented execution harness and runs the same infrastructure autonomously during benchmarking, enabling enterprise reliability to be evaluated under the controls intended for deployment.

\noindent\textbf{Computer-Use Agent Systems:}
Recent systems including Claude~\cite{anthropic2024computeruse} and OpenAI's~\cite{openai2025cua} computer-use capability, OpenCUA~\cite{opencua2025}, UI-TARS~\cite{uitars2025}, Holo3-35B-A3B~\cite{hai2025holo3modelfamily}, and Qwen3-VL~\cite{qwen3vl2025} paired with OmniParser~\cite{omniparser2025} demonstrate different approaches to screenshot-based computer interaction, ranging from native computer-use capabilities and GUI-specialized training to explicit visual element parsing~\cite{anthropic2024computeruse, openai2025cua, opencua2025, uitars2025, qwen3vl2025, omniparser2025, hai2025holo3modelfamily}. ERPBench evaluates representative systems from these approaches as benchmark subjects; its contribution is complementary to these model-development efforts, focusing on how reliably such agents perform enterprise operations rather than on improving GUI grounding itself.


Table~\ref{tab:comparison} positions ERPBench against existing benchmarks along observation modality, deployment, grounding, and grading. Of the four threads above, the benchmarks compared here fall into two: the general GUI thread (OSWorld~\cite{osworld2024}, Windows Agent Arena~\cite{waa2024}) and the enterprise computer-use evaluation thread (WorkArena~\cite{workarena2024}, SCUBA~\cite{scuba2025}, CRMArena-Pro~\cite{crmarena2025}, UI-CUBE~\cite{cristescu2025ui}); the agent safety and computer-use agent systems threads comprise safety studies and agent models rather than benchmarks. ERPBench advances the enterprise computer-use evaluation thread. 
ERPBench is the only benchmark that satisfies every row; each prior benchmark satisfies only a subset.


\section{ERPBench}
\label{sec:erpbench}
\begin{figure}[t!]
  \centering
  \includegraphics[width=\linewidth]{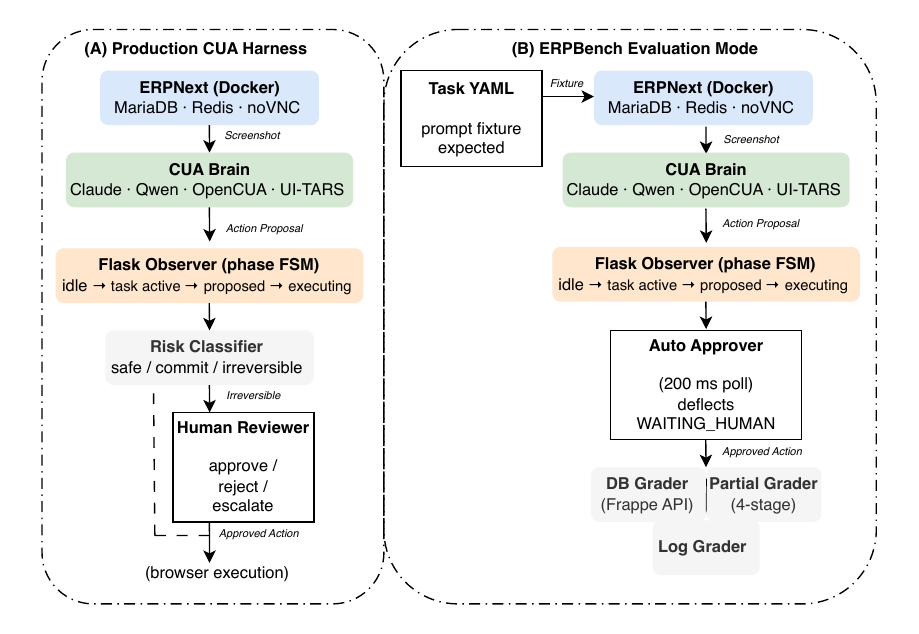}
  \caption{ERPBench system overview that compares the two harness configurations: \textbf{(A) Production CUA harness}. Demonstrates how proposed agent actions from different CUA brains transition through a Flask Finite State Machine (FSM) observer into a Risk Classifier. Risky or irreversible actions (such as deleting records) require explicit human approval before browser execution.
    \textbf{(B) Evaluation mode}. Replaces the human reviewer with an Auto Approver polling at 200 ms intervals. Post-execution state is processed by the DB Grader, Partial (Stage) Grader, and Log Grader.}
  \label{fig:pipeline}
\end{figure}

\vspace{-5pt}
\noindent ERPBench evaluates CUAs on ERPNext, an open-source ERP deployed locally in Docker (Fig.~\ref{fig:pipeline}).
The agent drives ERPNext through a browser as a person would: Chromium renders the application on a virtual display (Xvfb), streamed over noVNC.
At each step the agent observes only a $1280\times720$-pixel screenshot of this display and acts through simulated mouse and keyboard input at pixel coordinates.
No DOM or accessibility tree is exposed for perception or for control.
This pixel-in, coordinate-out channel mirrors how an agent must operate enterprise software behind remote-desktop or VDI deployments, where no API or accessibility hooks are available, the setting ERPBench is built to evaluate.
The rest of this section describes ERPBench's four components: the task suite (Sec.~\ref{ssec:tasks}), the harness that runs agents against ERPNext (Sec.~\ref{ssec:harness}), the graders that score each run (Sec.~\ref{ssec:grading}), and a human baseline (Sec.~\ref{ssec:human}).

\begin{figure}[ht]
  \centering
  \setlength{\fboxrule}{0.6pt}   
\setlength{\fboxsep}{0pt}      
    \fbox{\includegraphics[width=\columnwidth]{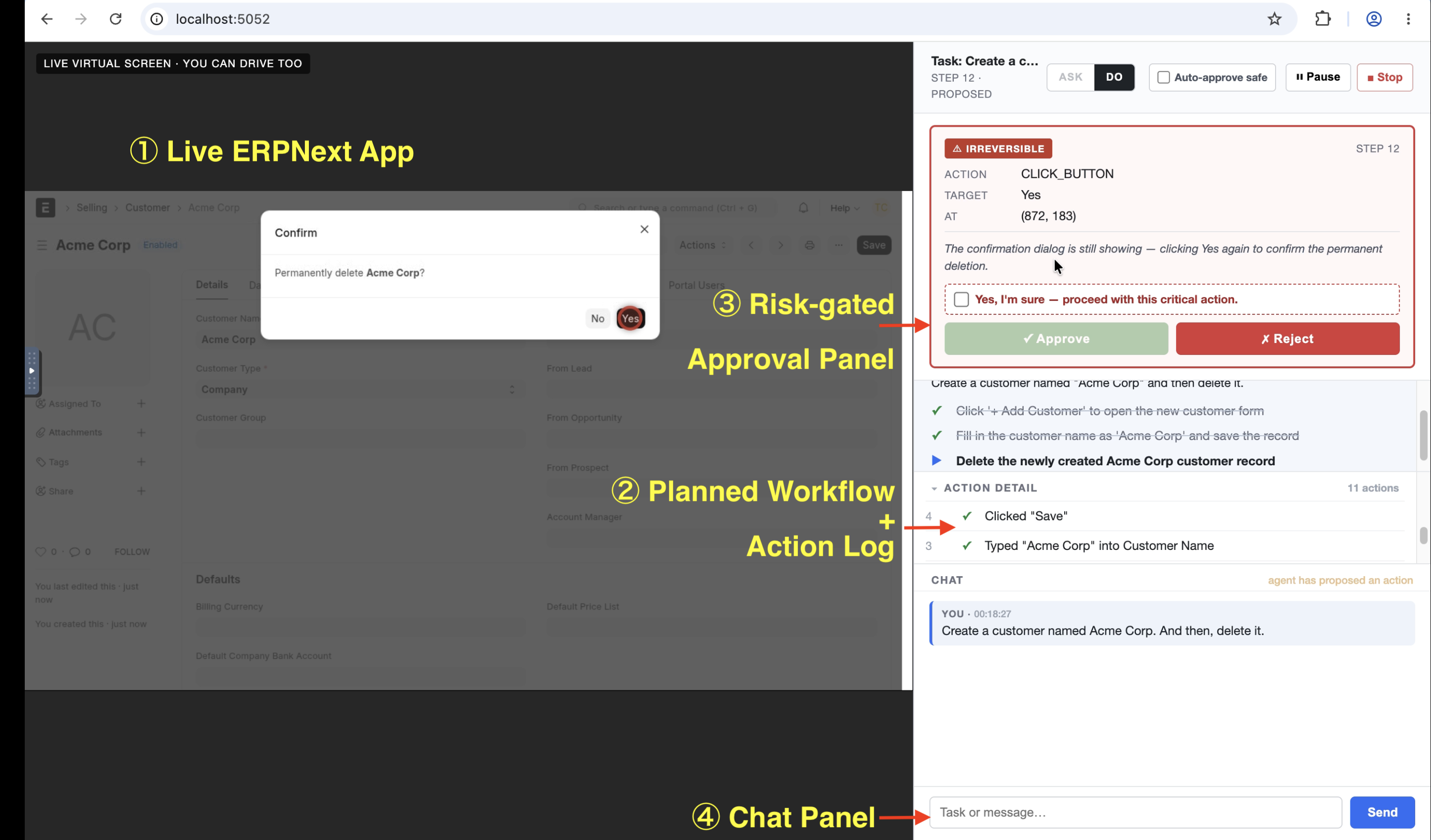}}
      \caption{ERPBench harness User Interface: Shows the live operator (human) interface divided into four functional panels: \textcircled{1} Live ERPNext App: The streaming screenshot-only workspace. \textcircled{2} Planned Workflow Action Log: Displays planned step sequences. \textcircled{3} Risk-Gated Approval Panel: Pauses execution on sensitive actions (e.g., irreversible deletion) pending operator review. \textcircled{4} Chat Panel: Provides a direct communication for operator intervention.}
      \vspace{-5pt}
  \label{fig:harness}
\end{figure}

\subsection{Task Design}
\label{ssec:tasks}
\vspace{-3pt}
ERPBench organizes tasks into three complexity tiers (T1--T3). 
T1 covers 20 single-record tasks in three types: \emph{text-edit} (typing into a text or long-text field, e.g., a customer's tax ID), \emph{select/toggle} (choosing via dropdown, autocomplete, checkbox, or date picker, e.g., a disabled flag), and \emph{create} (adding a simple record with a few fields, e.g., a new customer).
T2 introduces multi-field coordination and relational field selection, with 6-9 fields per record (e.g., creating a new Customer or Supplier from its full field set).
T3 introduces multi-screen chained workflows: the agent creates and links 2--4 dependent ERPNext documents in sequence (e.g., Lead $\to$ Opportunity $\to$ status advance, or Project $\to$ Task $\to$ Timesheet), and one variant additionally \emph{submits} a document, ERPNext's stronger and often irreversible commit, rather than only saving it.
Three T3 tasks are OSWorld-analog variants that start from a blank page, requiring the agent to navigate to ERPNext itself before any form work.
Because the harness and graders are specialized to ERPNext, running them unchanged on another benchmark is not straightforward; these variants instead reproduce the operating conditions of general GUI benchmarks inside ERPBench, providing a comparison point without a harness port.
Together, the tiers progressively increase interaction complexity, from a single field (T1) to a full multi-field record (T2) to a chain of linked records across screens (T3), while preserving a common correctness criterion: the final database state.
Each task is a YAML file specifying the prompt, a start URL, a fixture of pre-seeded fields, the expected field values, and grader configuration. Before each run, the fixture is seeded through the Frappe REST API under a UUID-suffixed record name (e.g., \texttt{Acme-Corp-RUN-edba5c7f}). This isolates runs without a full database reset.
\vspace{-7pt}

\subsection{Harness}
\label{ssec:harness}
\vspace{-4pt}
At each turn, the agent sees the current screenshot and proposes one action: move, scroll, click (field, button, link, or dropdown), type, select, set date, clear, press a key, or navigate. Each action carries target coordinates, a short reason, and a risk category (\emph{safe}, \emph{commit}, or \emph{irreversible}). If it is approved, the harness runs it as a real mouse or keyboard action and takes a new screenshot. The agent then proposes the next action.
The same harness runs in two modes. A Flask server tracks the task's state (idle $\to$ active $\to$ proposed $\to$ executing).
In deployment, actions tagged commit or irreversible (e.g., the save/submit bar) are surfaced for human approval before execution, while actions tagged safe execute directly (Fig.~\ref{fig:harness}). In evaluation, an auto-approver approves every action across all risk tiers, so a full task runs without a human.
Before each warm-start run, the browser is initialized at a task-specific start URL to remove URL-level navigation as a confound. Navigation within the ERP interface, from the start state to the target record or field, remains part of the agent's task and is scored as the first grading stage. Each run has a turn limit; a run that hits the limit is recorded as a recovery failure (see Sec.~\ref{ssec:failure} for additional details).

\subsection{Grading}
\label{ssec:grading}
\vspace{-4pt}
Each run is scored by three graders: Database grader, Stage grader, and Log grader. The database grader reads the target field values from ERPNext through the Frappe REST API after the run.
For a task with target fields $F$, let $y^*_i$ and $\hat{y}_i$ denote the expected and observed post-execution values of field $i\in F$. A per-field predicate $g(\hat{y}_i, y^*_i)\in\{0,1\}$ marks field $i$ correct under type-specific equivalence: numeric values agree within a tolerance $\tau$ (default $0.01$); Boolean-equivalent values agree (e.g., $1$ and \texttt{True}, or ``Yes'' and \texttt{True}); and strings are compared after HTML stripping and whitespace normalization. A task succeeds iff every target field is correct, i.e., $\prod_{i\in F} g(\hat{y}_i, y^*_i)=1$.

For multi-field T2 records, credit is field-weighted across the target fields rather than all-or-nothing. For multi-document T3 tasks, the grader is applied per document in the chain, and partial credit is the \emph{chain-depth}: the fraction of chain documents left correct in the database.
The stage grader awards partial credit over ordered stages, agent actions followed by a verification of state: \textbf{Navigation} (the target record or field is reached), \textbf{Interaction} (the required GUI manipulation is completed), \textbf{Commit} (a save or submit is invoked), and \textbf{Database} (the target fields match ground truth). Navigation, Commit, and Database apply to every tier; Interaction is specific to single-field edits, and chained workflows substitute chain-depth.

Each stage is evaluated independently from observable evidence, while the final database state remains the definitive criterion for task success. This localizes where a run failed, not just whether it failed.
The log grader reads the action and turn logs for action count, agent/API turns, latency, and, where available, token cost. Together, the three graders give each run a pass/fail result, a per-stage breakdown, and its cost.


\subsection{Human Baseline}
\label{ssec:human}
\vspace{-4pt}
As a human reference, three annotators independently completed all ERPBench tasks on the same ERPNext instance and were scored by the same database grader as the agents. Because humans operate the browser directly rather than through the screenshot-only agent interface, we report their task success and completion time and omit agent-specific metrics such as token usage. The reference deliberately spans different levels of platform familiarity: human annotators \#1 and \#2 are experienced ERPNext users and serve as an expert ceiling, while human annotator \#3 provides first-time-user reference, having never used ERPNext before.



\vspace{-5pt}

\section{Experiments and Results}
\label{sec:experiments}
\vspace{-5pt}

\begin{table*}[t!]\centering\scriptsize\setlength{\tabcolsep}{3pt}
\caption{ERPBench Benchmark results: Task Success and Efficiency Across Tiers. Demonstrates success rate (\%), mean GUI actions per run, 
mean duration per run (seconds), and mean input token usage per run across models.
Human rows are a human reference ceiling.}
\label{tab:main}
\begin{tabular*}{\textwidth}{@{\extracolsep{\fill}} l cccc cccc cccc @{}}
\toprule
 & \multicolumn{4}{c}{\textbf{T1}} & \multicolumn{4}{c}{\textbf{T2}} & \multicolumn{4}{c}{\textbf{T3}} \\
\cmidrule(lr){2-5}\cmidrule(lr){6-9}\cmidrule(lr){10-13}
\textbf{Model} & Success & Actions & Duration (s) & Input tokens (K)
 & Success & Actions & Duration (s) & Input tokens (K)
 & Success & Actions & Duration (s) & Input tokens (K) \\
\midrule
\multicolumn{13}{@{}l}{\textit{Proprietary}}\\
Claude Sonnet 4.6 & 94 & 11.7 & 89.2 & 231 & 100 & 25.6 & 170.1 & 784 & 100 & 36.2 & 322.7 & 2009 \\
\addlinespace
\multicolumn{13}{@{}l}{\textit{Open-weight}}\\
Holo3-35B-A3B & 34 & 6.5 & 37.1 & 34 & 0 & 14.2 & 58.6 & 93 & 3 & 6.8 & 56.0 & 40 \\
Qwen3-VL-32B & 32 & 4.9 & 29.1 & 76 & 0 & 7.7 & 39.4 & 107 & 3 & 10.5 & 61.9 & 149 \\
OpenCUA-32B  & 23 & 5.5 & 52.8 & 105 & 0 & 6.7 & 57.8 & 101 & 0 & 7.3 & 64.9 & 108 \\
UI-TARS-1.5-7B   & 9  & 13.1 & 59.2 & 51 & 0 & 26.2 & 99.0 & 103 & 0 & 25.3 & 111.4 & 107 \\
OpenCUA-7B   & 0  & 1.6 & 26.5 & 60 & 0 & 1.4 & 28.3 & 75 & 0 & 11.8 & 84.2 & 145 \\
\addlinespace
\multicolumn{13}{@{}l}{\textit{Human reference}}\\
Human \#1 & 100 & 7.5 & 41.4 & N/A & 95 & 25.1 & 85.7 & N/A & 100 & 27.1 & 99.9 & N/A \\
Human \#2 & 99 & 14.8 & 89.1 & N/A & 95 & 26.8 & 150.5 & N/A & 97 & 40.6 & 395.5 & N/A \\
Human \#3 & 96 & 8.9 & 42.3 & N/A & 90 & 24.6 & 93.2 & N/A & 87 & 28.7 & 100.7 & N/A \\
\bottomrule
\end{tabular*}
\end{table*}

We first describe the experimental setup, then report task success and efficiency, and finally analyze how and where the agents fail.

\subsection{Experimental Setup}
\label{ssec:setup}
\vspace{-3pt}
We evaluate six representative CUAs spanning proprietary and open-weight systems: Claude Sonnet 4.6, Qwen3-VL-32B, OpenCUA-32B, UI-TARS-1.5-7B, OpenCUA-7B, and Holo3-35B-A3B. Claude is accessed through the Anthropic API, while the open-weight models are self-hosted; Qwen3-VL-32B is paired with OmniParser for GUI grounding. Each task is executed for five independent runs under the same configuration and grader, yielding $100$ T1, $20$ T2, and $30$ T3 runs per model. Three human annotators complete the same tasks through the same database grader as a human reference.
On OSWorld-Verified these agents report 77.8 (Holo3-35B-A3B), 72.5 (Claude Sonnet 4.6), 34.8 (OpenCUA-32B), 27.4 (UI-TARS-1.5-7B), and 26.6 (OpenCUA-7B); Qwen3-VL-32B provides no native computer-use agent and is paired with OmniParser, so no OSWorld score applies.

\begin{table}[t]
\centering
\scriptsize
\setlength{\tabcolsep}{2pt}
\caption{Stage-wise localization within ERPBench illustrates where agents fail during execution. Demonstrates percent of runs reaching
each stage (Chain-depth as mean percent of the document chain completed).
Interaction is a T1-only stage (right value entered); Chain-depth is T3-only.
Agents reach the page but fail to persist correct database state. For T2, Database is field-weighted across the target fields, so it can exceed the all-or-nothing Success in Table~\ref{tab:main}.}
\label{tab:stages}
\begin{tabular*}{\columnwidth}{@{\extracolsep{\fill}} l cccc ccc cccc @{}}
\toprule
 & \multicolumn{4}{c}{\textbf{T1}} & \multicolumn{3}{c}{\textbf{T2}} & \multicolumn{4}{c}{\textbf{T3}} \\
\cmidrule(lr){2-5}\cmidrule(lr){6-8}\cmidrule(lr){9-12}
\textbf{Model}
 & \rotatebox{90}{Navigation} & \rotatebox{90}{Interaction} & \rotatebox{90}{Commit} & \rotatebox{90}{Database}
 & \rotatebox{90}{Navigation} & \rotatebox{90}{Commit} & \rotatebox{90}{Database}
 & \rotatebox{90}{Navigation} & \rotatebox{90}{Commit} & \rotatebox{90}{Chain-depth} & \rotatebox{90}{Database} \\
\midrule
\multicolumn{12}{@{}l}{\textit{Proprietary}}\\
Claude Sonnet 4.6 & 99 & 97 & 98 & 94 & 100 & 100 & 100 & 100 & 100 & 100 & 100 \\
\addlinespace
\multicolumn{12}{@{}l}{\textit{Open-weight}}\\
Holo3-35B-A3B & 90 & 53 & 47 & 34 & 100 & 5 & 0 & 33 & 17 & 4 & 3 \\
Qwen3-VL-32B & 90 & 30 & 42 & 32 & 100 & 15 & 5 & 67 & 23 & 6 & 3 \\
OpenCUA-32B  & 92 & 23 & 34 & 23 & 100 & 0  & 0 & 67 & 47 & 16 & 0 \\
UI-TARS-1.5-7B   & 95 & 30 & 68 & 9  & 100 & 85 & 3 & 60 & 80 & 3 & 0 \\
OpenCUA-7B   & 90 & 0  & 3  & 0  & 100 & 10 & 0 & 33 & 3  & 0  & 0 \\
\addlinespace
\multicolumn{12}{@{}l}{\textit{Human reference}}\\
Human \#1 & 100 & 100 & 100 & 100 & 100 & 100 & 95 & 100 & 100 & 100 & 100 \\
Human \#2 & 100 & 97  & 100 & 99  & 100 & 100 & 99 & 100 & 100 & 98 & 97 \\
Human \#3 & 100 & 93 & 98 & 96 & 100 & 100 & 94 & 100 & 100 & 94 & 87 \\
\bottomrule
\end{tabular*}
\end{table}

\begin{table}[t]
\centering
\scriptsize
\setlength{\tabcolsep}{2pt}
\caption{Additional efficiency metrics across tiers. Demonstrates agent/API turn counts,
generated output tokens, and per-turn median thinking times (seconds).}
\label{tab:efficiency-detail}
\begin{tabular*}{\columnwidth}{@{\extracolsep{\fill}} l ccc ccc ccc @{}}
\toprule
 & \multicolumn{3}{c}{\textbf{T1}} & \multicolumn{3}{c}{\textbf{T2}} & \multicolumn{3}{c}{\textbf{T3}} \\
\cmidrule(lr){2-4}\cmidrule(lr){5-7}\cmidrule(lr){8-10}
\textbf{Model}
 & \rotatebox{90}{Turns} & \rotatebox{90}{Output tokens} & \rotatebox{90}{Thinking (s)}
 & \rotatebox{90}{Turns} & \rotatebox{90}{Output tokens} & \rotatebox{90}{Thinking (s)}
 & \rotatebox{90}{Turns} & \rotatebox{90}{Output tokens} & \rotatebox{90}{Thinking (s)} \\
\midrule
\multicolumn{10}{@{}l}{\textit{Proprietary}}\\
Claude Sonnet 4.6 & 13.8 & 2{,}255 & 4.320 & 27.6 & 4{,}455 & 4.400 & 38.2 & 6{,}468 & 6.217 \\
\addlinespace
\multicolumn{10}{@{}l}{\textit{Open-weight}}\\
Holo3-35B-A3B & 8.5 & 1{,}634 & 2.095 & 15.2 & 2{,}866 & 2.000 & 9.7 & 2{,}054 & 2.067 \\
Qwen3-VL-32B & 8.4 & 543 & 1.212 & 10.9 & 687 & 1.536 & 14.5 & 1{,}068 & 1.554 \\
OpenCUA-32B & 11.9 & 751 & 2.660 & 11.1 & 684 & 3.009 & 11.4 & 775 & 2.527 \\
UI-TARS-1.5-7B & 15.8 & 1{,}503 & 1.875 & 30.0 & 2{,}871 & 1.850 & 29.4 & 2{,}825 & 1.817 \\
OpenCUA-7B & 6.9 & 495 & 1.471 & 8.3 & 548 & 1.492 & 14.8 & 1{,}072 & 1.623 \\
\bottomrule
\end{tabular*}
\end{table}


\begin{table}[t]
\centering
\scriptsize
\setlength{\tabcolsep}{3pt}
\caption{
Partial credit across tiers by sub-task type (\%). Text-edit and Select/Toggle (T1) and
Multi-field (T2) report the mean partial score, the fraction of that tier's
grading stages passed, averaged over tasks and runs. Create (T1) reports the
pass rate for simple single-record creates (graded pass/fail). T3 reports the
mean chain-depth (fraction of the document chain completed), split into
warm-start and blank-start (OSWorld-analog) variants.
}
\label{tab:partial}
\begin{tabular*}{\columnwidth}{@{\extracolsep{\fill}} l ccc c cc @{}}
\toprule
 & \multicolumn{3}{c}{\textbf{T1}} & \textbf{T2} & \multicolumn{2}{c}{\textbf{T3}} \\
\cmidrule(lr){2-4}\cmidrule(lr){5-5}\cmidrule(lr){6-7}
\textbf{Model} & \shortstack{Text-\\edit} & \shortstack{Select/\\Toggle}
 & Create & \shortstack{Multi-\\field} & \shortstack{Warm-\\start}
 & \shortstack{Blank-\\start} \\
\midrule
Claude Sonnet 4.6 & 96 & 97 & 100 & 100 & 100 & 100 \\
Holo3-35B-A3B & 61 & 58 & 15 & 35 & 9 & 0 \\
Qwen3-VL-32B & 43 & 49 & 70 & 40 & 11 & 2 \\
OpenCUA-32B & 37 & 49 & 30 & 33 & 22 & 9 \\
UI-TARS-1.5-7B & 47 & 54 & 35 & 63 & 7 & 0 \\
OpenCUA-7B & 22 & 31 & 0 & 37 & 0 & 0 \\
\bottomrule
\end{tabular*}
\end{table}

Table~\ref{tab:main} reports task success and efficiency across the three tiers; Table~\ref{tab:stages} decomposes each run into ordered stages and illustrates where agents fail during execution. 
Success is the share of runs in which every target field matches the database; Actions, Duration, and Input tokens are per-run means. Actions counts executed GUI operations, while Turns (Table~\ref{tab:efficiency-detail}) counts agent-model calls and exceeds Actions when a turn yields no executed operation (e.g., a rejected or terminal step).
The stage columns report the share of runs reaching each stage: Navigation (the target record or field is reached), Interaction (the correct value is entered; T1 only), Commit (a save or submit is invoked), Database (the target fields match ground truth; field-weighted for T2), and, for T3, Chain-depth (the mean fraction of the document chain left correct). Human rows are a human reference ceiling and omit token metrics, which are agent-specific.

\subsection{Results}
\label{ssec:results}
\vspace{-5pt}
Strong general GUI performance does not translate into reliable enterprise execution. 
Claude reaches 94\%, 100\%, and 100\% success across the three tiers, approaching the human reference; however, its token requirements scale heavily, from 231K tokens per run on T1 to roughly 2.0M on T3. 
By contrast, the strongest open models, Holo3-35B-A3B and Qwen3-VL-32B, reach only 34\% and 32\% on T1 and fall to 0--3\% on T2 and T3, while OpenCUA-7B fails every task (Table~\ref{tab:main}).
General GUI scores do not predict enterprise reliability: Holo3-35B-A3B outscores Claude on OSWorld-Verified (77.8 against 72.5) yet reaches 34\% on T1 where Claude reaches 94\%.

Table~\ref{tab:stages} localizes where this competence leaks away. Open models successfully navigate to target records in 90--95\% of T1 runs, but fail at interaction and database persistence. Their interaction rate falls to 0--53\% and their database-correct rate to 0--34\%: they arrive at the right screen yet cannot reliably enter and persist the intended value.
UI-TARS is the clearest case: it reaches the target in 95\% of runs and fires a save in 68\%, yet only 9\% leave the correct value in the database, so nearly every save it commits is wrong. 
The pattern sharpens with tier depth: on T2 the open models reach the record in every run yet are database-correct in at most 5\%, with UI-TARS committing in 85\% of runs but database-correct in 3\%, and on T3 the open models complete no chain end to end.
Claude, by contrast, stays near ceiling across every stage (T1 navigate 99\%, interact 97\%, commit 98\%, database 94\%).
The open models' collapse with tier depth mirrors the capability cliff UI-CUBE~\cite{cristescu2025ui} reports as enterprise tasks grow more complex, and the broader enterprise gap EntWorld~\cite{mo2026entworld} observes. ERPBench’s stage-wise, database-level grading provides a finer resolution on this gap by isolating runs where an agent navigates and even commits yet leaves the wrong value in the record, a silent failure that screenshot-level or interaction-level grading would misclassify as success.

Table~\ref{tab:main} also shows the cost of the longer horizon. Claude completes all six T3 chained workflows (30/30), matching the best human annotator, but at 36 actions and 323\,s per run versus 12 actions and 89\,s on T1.
The open models stay cheap only because they stop early: OpenCUA-7B averages 1.6 actions on T1 and Qwen3-VL 4.9, reflecting premature loop termination rather than task efficiency.


Table~\ref{tab:efficiency-detail} reports the remaining per-run metrics; they track the same pattern, Claude's turn counts and output tokens grow with tier depth (13.8 turns in T1 to 38.2 turns in T3, 2.3K output tokens in T1 to 6.5K output tokens in T3) while its per-turn thinking time grows more modestly, from about 4\,s on T1--T2 to 6\,s on T3.
Open models exhibit low average turns and low output token counts primarily because they hit loop-breaks or terminate early.
Table~\ref{tab:partial} shows partial credit for all tiers and analyzes performance across specific sub-task types. 
With the exception of Holo3, open models perform worse on exact free-text field entry (22--47\%) than on select/toggle choices (31--54\%), i.e., typing an exact value is harder than picking a known option.  
In T3, removing the navigation crutch (starting URL prompts) causes chain completion depth to drop significantly. For example, OpenCUA-32B drops from 22\% warm-start depth to 9\% blank-start depth.

For human annotators, familiarity has only a modest effect: the experts reach 99--100\% on T1, 95\% on T2, and 97--100\% on T3, and the first-time user still reaches 96\%, 90\%, and 87\% respectively. Even this first-time user far exceeds every open model (which top out at 34\% on T1 and 3\% on T3), leaving Claude as the only agent that performs in the human range; ERPBench's difficulty for agents is therefore not only a matter of ERP-specific expertise. Across the 24 T1 and T2 tasks, the three annotators reached identical outcomes on 16 (67\%) and agreed to within a single run on 23 (96\%), with the first-time user accounting for most of the small differences.

\subsection{Failure Analysis}
\label{ssec:failure}
\begin{figure}[ht]
  \centering
  \includegraphics[width=\linewidth]{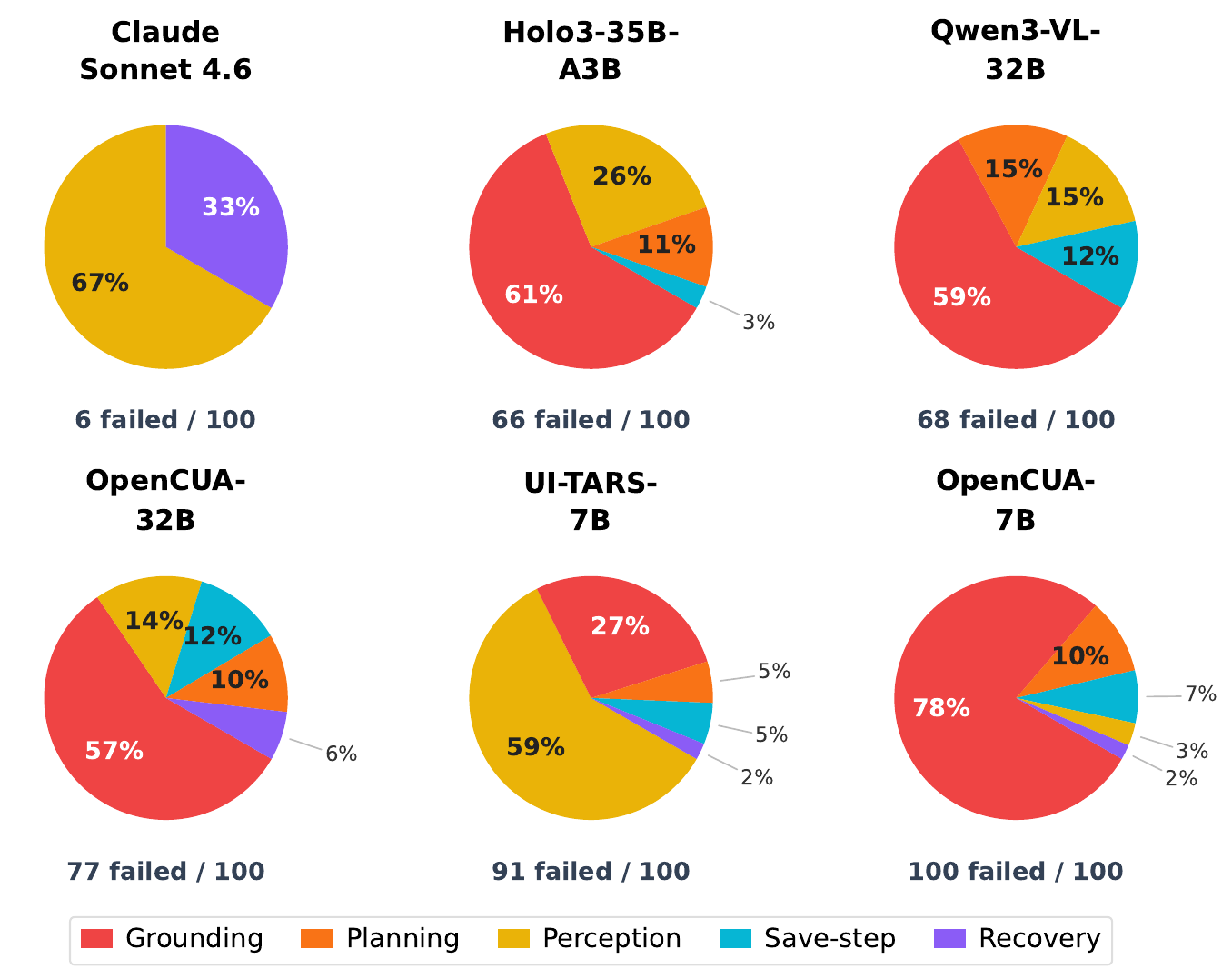}
  \caption{Failure taxonomy over failed T1 runs, by model. Most failures occur after successful navigation, clustering in grounding, save-step, and perception rather than planning: reaching the right screen does not imply completing the task correctly.}
  \label{fig:failures}
\end{figure}

\vspace{-5pt}
Figure~\ref{fig:failures} indicates the failure taxonomy over failed T1 runs. It categorizes non-successful T1 runs into five hierarchical modes. \emph{Planning} failures never reach the target record. \emph{Grounding} failures reach the record but act on the wrong element, so the value is never entered. \emph{Save-step} failures enter the target value correctly but fail to trigger save. \emph{Perception} failures save successfully, but an incorrect value is written to the persistent database (silent failure). \emph{Recovery} failures get trapped in interaction loops or exhaust the turn budget without making progress.

Open-weight models fail predominantly post-navigation rather than in planning. For instance, of Qwen3-VL-32B's 68 failed T1 runs, 40 are Grounding, 10 Planning, 10 Perception, and 8 Save-step. Of UI-TARS-1.5-7B's 91 failures, 54 are Perception and 25 Grounding.

Each failed run is assigned to a single mode using the priority order Recovery, Planning, Perception, Save-step, then Grounding. Recovery is checked first because a run that loops or times out can do so at any stage. Planning is next (never reaching the record), followed by the two silent-failure checks, Perception (a wrong value was saved) and Save-step (a correct value was never saved), leaving Grounding as the residual category for runs that reach the record but act on the wrong element.

Two of these modes are exactly why enterprise evaluation needs database-level grading. High visual competence on screen does not correlate directly with database accuracy. In Save-step and Perception failures the agent acts, the screen shows a successful save, yet the persistent record is wrong or unchanged. 
Evaluating computer-use agents solely through screenshot comparisons or execution traces hides silent enterprise commit failures, underscoring the necessity of database-grounded verification.
These silent failures do not surface on screen but propagate into the business records that downstream processes depend on, which is precisely what state-grounded evaluation is designed to catch.

Stage-wise grading and the failure taxonomy are complementary lenses: the stages identify \emph{where} in the execution pipeline a run breaks, while the taxonomy characterizes \emph{how} it breaks.

\vspace{-5pt}

\section{Conclusion}
\label{sec:conclusion}
\vspace{-5pt}
We introduced ERPBench, a benchmark for screenshot-only computer-use agents on a real, open, self-hosted ERP. 
Tasks are graded against the database state the agents leave behind, and stage-wise evaluation identifies where failures occur across navigation, interaction, commit, and database correctness.
We also contribute the production-grade harness used to run the agents, which gates actions behind human approval in deployment, and runs the same harness autonomously for evaluation.

Across six agents, strong general GUI performance did not carry over to enterprise work. Models that navigate well still fail at interaction and commit, and some appear to succeed on screen while leaving the database wrong or unchanged. Interaction-level evaluation can therefore overestimate enterprise reliability; database-grounded evaluation reveals failures that remain invisible at the GUI level. ERPBench currently covers single-field edits, multi-field record creation and multi-screen chained workflows. 
We intend to release the benchmark and its evaluation artifacts.

Because these failures localize to interaction, commit, and persistence, they point targeted improvement, such as fine-tuning on state-grounded trajectories, at those stages, while the human-in-the-loop harness makes pixel-only agents deployable in the meantime.

\section{Acknowledgments}
The authors are grateful to Eduardo Salamanca de Diego and Jordan Ackerman for their guidance and continued support of this work, and to Christopher Clarke, Shafiuddin Rehan Ahmed, and Teja Kanchinadam for their generous assistance with model deployment and debugging. The authors also thank Shanka Subhra Mondal and Dhaval Potdar for their helpful review of the manuscript.

The authors used Claude (Anthropic) for language refinement and structural organization of this paper. Generative AI tools were not used to generate experiments, perform data analysis, or produce scientific claims. All methodologies, results, and conclusions were developed independently by the authors.

\section{Compliance with Ethical Standards}
The human reference was produced by the authors themselves; no external participants and no sensitive personal data were involved, and no institutional review board approval was required.

\vfill\pagebreak

\label{sec:refs}



\bibliographystyle{IEEEbib}
\bibliography{strings,refs}

\begin{thebibliography}{10}

\bibitem{uitars2025}
Yujia Qin et~al.,
\newblock ``{UI-TARS: Pioneering Automated {GUI} Interaction with Native Agents},''
\newblock {\em arXiv:2501.12326}, January 2025.

\bibitem{anthropic2024computeruse}
{Claude Platform Docs, Anthropic},
\newblock ``Computer use tool,'' \url{https://platform.claude.com/docs/en/agents-and-tools/tool-use/computer-use-tool}, November 2024.

\bibitem{openai2025cua}
{OpenAI Developers, OpenAI},
\newblock ``Computer use,'' \url{https://developers.openai.com/api/docs/guides/tools-computer-use}, 2025.

\bibitem{opencua2025}
Xinyuan Wang et~al.,
\newblock ``Opencua: Open foundations for computer-use agents,''
\newblock {\em arXiv:2508.09123}, October 2025.

\bibitem{osworld2024}
Tianbao Xie et~al.,
\newblock ``{OSWorld: Benchmarking Multimodal Agents for Open-Ended Tasks in Real Computer Environments},''
\newblock {\em Proceedings of the Advances in Neural Information Processing Systems}, vol. 37, pp. 52040--52094, December 2024,
\newblock {Vancouver, Canada}.

\bibitem{waa2024}
Rogerio Bonatti, Dan Zhao, Francesco Bonacci, Dillon Dupont, Sara Abdali, Yinheng Li, Yadong Lu, Justin Wagle, Kazuhito Koishida, Arthur Bucker, Lawrence Jang, and Zheng Hui,
\newblock ``{Windows Agent Arena: Evaluating Multi-Modal OS Agents at Scale},''
\newblock {\em Proceedings of the International Conference on Machine Learning}, pp. 4874--4910, July 2025,
\newblock {Vancouver, Canada}.

\bibitem{zhou2024webarena}
Shuyan Zhou, Frank~F. Xu, Hao Zhu, Xuhui Zhou, Robert Lo, Abishek Sridhar, Xianyi Cheng, Tianyue Ou, Yonatan Bisk, Daniel Fried, Uri Alon, and Graham Neubig,
\newblock ``{WebArena: A Realistic Web Environment for Building Autonomous Agents},''
\newblock {\em Proceedings of the International Conference on Learning Representations}, pp. 15585--15606, May 2024,
\newblock {Vienna, Austria}.

\bibitem{scuba2025}
Yutong Dai, Krithika Ramakrishnan, Jing Gu, Matthew Fernandez, Yanqi Luo, Viraj Prabhu, Zhenyu Hu, Silvio Savarese, Caiming Xiong, Zeyuan Chen, and Ran Xu,
\newblock ``{SCUBA: Salesforce Computer Use Benchmark},''
\newblock {\em Proceedings of the International Conference on Learning Representations}, pp. 24363--24386, April 2026,
\newblock Rio de Janeiro, Brazil.

\bibitem{crmarena2025}
Kung-Hsiang Huang, Akshara Prabhakar, Onkar Thorat, Divyansh Agarwal, Prafulla~Kumar Choubey, Yixin Mao, Silvio Savarese, Caiming Xiong, and Chien-Sheng Wu,
\newblock ``{CRMArena-Pro: Holistic Assessment of LLM Agents Across Diverse Business Scenarios and Interactions},''
\newblock {\em Transactions on Machine Learning Research}, p.~35, January 2026.

\bibitem{cristescu2025ui}
Horia Cristescu, Charles Park, Trong~Canh Nguyen, Sergiu Talmacel, Alexandru-Gabriel Ilie, and Stefan Adam,
\newblock ``{UI-CUBE: Enterprise-Grade Computer Use Agent Benchmarking Beyond Task Accuracy to Operational Reliability},''
\newblock {\em arXiv:2511.17131}, November 2025.

\bibitem{erpnext}
{Frappe},
\newblock ``{ERPNext: Free and Open Source Enterprise Resource Planning},'' \url{https://github.com/frappe/erpnext}, 2026.

\bibitem{qwen3vl2025}
Shuai Bai, Yuxuan Cai, Ruizhe Chen, et~al.,
\newblock ``{Qwen3-VL} technical report,''
\newblock {\em arXiv:2511.21631}, November 2025.

\bibitem{hai2025holo3modelfamily}
{H Company},
\newblock ``{Holo3 - Open Foundation Models for Navigation and Computer Use Agents},'' \url{https://huggingface.co/Hcompany/Holo3-35B-A3B}, 2026.

\bibitem{workarena2024}
Alexandre Drouin, Maxime Gasse, Massimo Caccia, Issam~H. Laradji, Manuel Del~Verme, Tom Marty, David Vazquez, Nicolas Chapados, and Alexandre Lacoste,
\newblock ``{WorkArena: How Capable are Web Agents at Solving Common Knowledge Work Tasks?},''
\newblock {\em Proceedings of the International Conference on Machine Learning}, vol. 235, pp. 11642--11662, July 2024,
\newblock {Vienna, Austria}.

\bibitem{mo2026entworld}
Ying Mo, Yu~Bai, Dapeng Sun, Yuqian Shi, Yukai Miao, Li~Chen, and Dan Li,
\newblock ``{EntWorld: A Holistic Environment and Benchmark for Verifiable Enterprise GUI Agents},''
\newblock {\em arXiv:2601.17722}, January 2026.

\bibitem{turan2026oversightcapacitycalibratingagent}
Emre Turan,
\newblock ``{Oversight Has a Capacity: Calibrating Agent Guards to a Subjective, Fatiguing Human},''
\newblock {\em arXiv:2606.08919}, June 2026.

\bibitem{wang2026reframingllmagentsecurity}
Peiran Wang, Ying Li, and Yuan Tian,
\newblock ``{Reframing LLM Agent Security as an Agent-Human Interaction Problem},''
\newblock {\em arXiv:2605.24309}, May 2026.

\bibitem{chen2026lpsbenchbenchmarkingsafetyawareness}
Tianyu Chen, Chujia Hu, Ge~Gao, Dongrui Liu, Xia Hu, and Wenjie Wang,
\newblock ``{LPS-Bench: Benchmarking Safety Awareness of Computer-Use Agents in Long-Horizon Planning under Benign and Adversarial Scenarios},''
\newblock {\em arXiv:2602.03255}, February 2026.

\bibitem{zhang2026invisibleinkthreatsadversarial}
Jia-Chen Zhang, Ze-Yu Zhang, and Kai-Wei Zhang,
\newblock ``{Invisible Ink Threats: Adversarial Goals Behind Legitimate Tasks in Computer-Use Agents},''
\newblock {\em arXiv:2608.02018}, August 2026.

\bibitem{omniparser2025}
Yadong Lu, Jianwei Yang, Yelong Shen, and Ahmed Awadallah,
\newblock ``{OmniParser for Pure Vision Based GUI Agent},''
\newblock {\em arXiv:2408.00203}, August 2024.

\end{thebibliography}

\end{document}